\documentclass[a4paper]{article}
\usepackage{spconf}
\usepackage{times}
\usepackage{url}
\usepackage{latexsym}
\usepackage{titlesec}
\usepackage{bm}
\usepackage{amssymb}
\usepackage{epsfig}
\usepackage{amsmath}
\usepackage{multirow}
\usepackage{graphicx}
\usepackage{stmaryrd}
\usepackage[usenames, dvipsnames]{color}
\usepackage{float}
\usepackage{caption}
\usepackage[utf8x]{inputenc}
\usepackage[titletoc,toc,title]{appendix}

\def\reg{{\rm\ooalign{\hfil
     \raise.07ex\hbox{\scriptsize R}\hfil\crcr\mathhexbox20D}}}
\allowdisplaybreaks
\usepackage{graphics}

\makeatletter
\newcommand\footnoteref[1]{\protected@xdef\@thefnmark{\ref{#1}}\@footnotemark}
\makeatother

\newcommand{\todo}[1]{\textcolor{WildStrawberry}{\bf\small TODO}}

\title{Towards Fluent Translations from Disfluent Speech}
\name{Elizabeth Salesky$^1$, Susanne Burger$^1$, Jan Niehues$^2$, and Alex Waibel$^{1,2}$}
\address{
  $^1$Carnegie Mellon University, Pittsburgh PA, U.S.A.\\
  $^2$Karlsruhe Institute of Technology, Karlsruhe, Germany\\
  \texttt{esalesky@cs.cmu.edu}
  }

\begin{document}
\maketitle

\begin{abstract}
When translating from speech, special consideration for conversational speech phenomena such as disfluencies is necessary.
Most machine translation training data consists of well-formed written texts, causing issues when translating spontaneous speech.
Previous work has introduced an intermediate step between speech recognition (ASR) and machine translation (MT) to remove disfluencies, making the data better-matched to typical translation text and significantly improving performance.
However, with the rise of end-to-end speech translation systems, this intermediate step must be incorporated into the sequence-to-sequence architecture. 
Further, though translated speech datasets exist, they are typically news or rehearsed speech without many disfluencies (e.g. TED), or the disfluencies are translated into the references (e.g. Fisher). 
To generate clean translations from disfluent speech, cleaned references are necessary for evaluation.
We introduce a corpus of cleaned target data for the Fisher Spanish-English dataset for this task. 
We compare how different architectures handle disfluencies and provide a baseline for removing disfluencies in end-to-end translation.
\end{abstract}


\section{Introduction}
\label{sec:intro}

Spoken language translation applications suffer due to disfluencies in spontaneous speech.
In conversational speech, speakers often use disfluencies such as filler words, repetitions, false starts, and corrections. 
These speech phenomena interfere with recognition and translation steps.
In this work, we use disfluent conversational speech from the Fisher Spanish dataset\footnote{\label{ldcfisher}LDC2010S01 and LDC2010T04} which has been translated to English \cite{post2013improved} with the disfluencies faithfully translated. 

Machine translation systems are typically trained using well-structured and cleanly written text. 
The mismatch between clean training data and test data with speech phenomena causes a drop in performance. 
Systems to detect and remove disfluencies from input speech, creating cleaner source data for MT, have been shown to greatly improve the performance of spoken language translation systems, even on broadcast news and TED talks where these phenomena are less common \cite{cho2013crf,cho2014lrec,wang2010disfluency,honal2005spkdisfluencies,zayats2016disfluency}. 

Before end-to-end models, different datasets could be used to train the speech recognition and translation components of a speech translation system.
Where aligned speech and translations exist, the data is usually clean speech$\shortrightarrow$clean text, as in news data or TED talks, or disfluent speech$\shortrightarrow$disfluent translations, as in Fisher or meeting data, where the disfluencies have been faithfully included in the references for completeness, but are not labeled. 
Some corpora with labeled disfluencies exist; these labeled sections can be removed to create clean target text. However, only parts of these corpora have been translated and/or released \cite{cho2014lrec,burger2002isl}.
Using disfluent/disfluent parallel data, we cannot score generated translations with disfluencies removed; we need cleaned reference translations.
We used MTurk to create cleaned reference translations for the Fisher Spanish-English data. 
This data is being prepared for public release.\footnote{The link to the data will be placed here when the data is released.}

Disfluency recognition and removal has previously been performed as an intermediate step between speech recognition (ASR) and machine translation (MT), to make transcripts more similar to typical machine translation data. 
With the rise of end-to-end sequence-to-sequence speech translation systems \cite{weiss2017sequence}, disfluency removal would need to be incorporated into the model instead of handled as a separate step. 
Further, implicit handling in the model architecture may promote the ability to recognize disfluencies and corrections specific to current data, outside of a set of handcrafted labels. 
We hope this data will promote further research in this area.


\section{Data}

For our experiments, we use the Fisher Spanish dataset\footnoteref{ldcfisher}, composed of telephone conversations between mostly native Spanish speakers.
The corpus consists of 819 transcribed conversations on provided topics between strangers, yielding ${\sim}160$ hours of speech and 150k utterances. 
The transcripts were translated and released by JHU\footnote{joshua-decoder.org/fisher-callhome-corpus} \cite{post2013improved}, with one reference for the training data and four each for the dev and test sets. 

This data is conversational and disfluent. 
Disfluencies can be filler words and hesitations, discourse markers (\textit{you know}, \textit{well}, \textit{mm}), repetitions, corrections and false starts, among others.
The reference translations maintain and translate where possible the disfluencies in the Spanish source.
Examples of certain types of disfluencies shown in both source and target are below in Table \ref{example disfluencies}.

\begin{table}[ht]
\centering
\begin{tabular}{ll}
\bf Hesitation  & eh, eh, eh, um, yo pienso que es así. \\
                & uh, uh, uh, um, i think it's like that. \\
\bf Repetition  & Y, y no cree que, que, que, \\
                & And, and I don't believe that, that, that \\
\bf Correction  & no, no puede, no puedo irme para ... \\
                & no, it cannot, I cannot go there ... \\
\bf False start & porque qué va, mja ya te acuerda que ... \\
                & because what is, mhm do you recall now that ... \\
                
\end{tabular}
\caption{Examples of disfluencies in Fisher Spanish-English, in the Spanish transcripts and English reference translations}
\label{example disfluencies}
\end{table}

We note that there can be many different and often overlapping types of disfluencies in a single utterance, as in this example from the Spanish data, `también tengo um eh estoy tomando una una clase ...' which contains filler words (\textit{um, eh}), a correction (\textit{tengo $\rightarrow$ estoy tomando}), and repetition (\textit{una, una}) with overlapping scope. 
Additionally, context affects whether some word types are disfluent (\textit{so, oh, ...}), and so removing them in all cases will affect meaning.
Disfluency removal is a more complex problem than merely recognizing disfluencies from a compiled list. 

\subsection{Mechanical Turk}

To create clean `copy-edited' reference translations, we crowd-sourced the task on Amazon Mechanical Turk. 
Turkers were presented with the original English translations in context and asked to remove certain types of disfluencies while maintaining meaning. 
We specify here filler words, repetitions, corrections, false starts, with examples of each. 
Where an utterance was deemed to have only disfluencies, Turkers were instructed to enter `None' for no content and specify why in the comments.
Turkers were paid a competitive rate equating to a U.S. hourly minimum wage.
Each ‘Human Intelligence Task’ (HIT) was limited to 5 utterances to not overwhelm Turkers. 
The first utterance was a control, allowing Turkers to familiarize themselves with the task, the results of which are not included in our data.
Utterances with only 1 token were not crowd-sourced but labeled manually by us.

We required that Turkers had an approval rate of 95\%.
We had 1250 unique workers complete 26,270 HITS. 
The ratio of approved to rejected HITs was 25:1.
HITs were rejected if answered implausibly fast, the answers were incomplete, or
they included clearly unrelated content. 
Each utterance in the dev and test sets was cleaned by at least 2 Turkers to reduce variability, and for the larger training set, one Turker.

\subsection{Original vs Cleaned References}

The cleaned references contain on average 3 fewer tokens per sentence, reducing the average sentence length from 11.3 to 8.2. 
Most utterances contained at least one disfluency; only 35\% of utterances were unchanged by Turkers. 
However, 16,829 sentences or 10.5\% of all utterances were marked only disfluencies. 
These ranged from single token utterances (`Mhm') to potentially several (`Hmm mm hmm mm we').
They were typically very short (fewer than three tokens) and contained only filler words or false starts; in context, most can be viewed as backchanneling. 
Backchanneling, or verbal cues to indicate attention, can sometimes convey information or meaningful reactions (`oh?'), and other times, may be classified as disfluencies. 
To determine which, it is important to view the utterance in context.
In most cases, Turkers reduced these utterances to `None'. 
Below we discuss annotator agreement, and sources of least disagreement. 

Below is an example of the types of changes made by Turkers. 
We show the original Spanish transcript and the disfluent English translation, along with the generated clean reference. 
We use NIST's sclite tool \cite{fiscus1998sclite} to evaluate changes made by Turkers in terms of insertions (I), deletions (D), and substitutions (S). 
\begin{table}[ht]
\centering
\setlength\tabcolsep{0.75pt} 
\begin{tabular}{lcccccccccc}
\bf SRC & Y, & bueno, & y & que, & aunque & no & se & ve & \\ \hline
\bf REF & and, & well, & and & that, & even & though & you & don't & see & him \\
\bf CLEAN & *** & *** & and & *** & even & though & you & don't & see & him \\ 
\bf Eval: & D & D &  & D &  &  &  &  &  \\
\end{tabular}
\caption*{Example of generated cleaned references (CLEAN) with original Spanish source (SRC) and disfluent English target (REF).}

\label{ISD example}
\end{table}

While many disfluencies can be removed through deletions, false starts and corrections can often lead to insertions, substitutions, or reorderings in the cleaned text. 
Table \ref{isd} shows the percentage breakdowns of insertions, deletions, and substitutions made by Turkers in cleaning each of the datasets.

\begin{table}[ht]
\centering
\begin{tabular}{lccc} \hline
\bf Dataset & Insertions & Deletions & Substitutions \\ \hline 
\bf train   & 0.6\% & 25.8\% & 2.8\%  \\
\bf dev     & 1.5\% & 31.7\% & 5.2\%  \\
\bf dev2    & 1.0\% & 33.2\% & 4.5\%  \\
\bf test    & 1.2\% & 31.4\% & 4.5\%  \\ \hline
\end{tabular}
\caption{Percentages of token insertions, deletions, and substitutions made by Turkers in generating the cleaned reference translations.}
\label{isd}
\end{table}

To verify the content changed by Turkers, we first look at the agreement between Turkers.
For the \texttt{dev}, \texttt{dev2}, and \texttt{test} datasets, we collected annotations for each utterance from two different Turkers to measure consistency. 
We use the two collected annotations for each utterance to make two clean reference translations. 
We can look at the BLEU (n-gram precision) \cite{papineni2002bleu} between the two references as a measure of annotator agreement, shown in Table \ref{annotator agreement bleu}.
The typical preprocessing scheme for this dataset is lowercased and with all punctuation removed \cite{post2013improved,kumar2014some,weiss2017sequence}; to provide a fair comparison, we remove punctuation and case to test annotator agreement. 
We show the average BLEU against a single reference, as a multi-reference score would not be a fair comparison here.
We find a high level of agreement between MTurk annotators, suggesting this is a task that can be crowd-sourced.
For context, we additionally show the average BLEU score between the pairs of the four original references for each of these datasets, as well as the BLEU score of the two clean references against the original disfluent translations.

We find that the inter-annotator BLEU score is very high across the cleaned corpus, and considerably higher than the original data's inter-annotator BLEU. 
This is unsurprising, as in our task, the two Turkers are given the same English sentence and told to edit specific content; unaltered content will be the same between the two Turked references. 
In the original collection, the four Turkers were given the same Spanish source sentence and independently generated translations, leading to more variability.
The original inter-annotator BLEU can serve as a benchmark for our translation systems, as this is the BLEU between human translators on this data.

We also compare the clean translations to the original disfluent translations. 
Annotator-Original uses the original data as references to score the new clean MTurk data as `hypotheses', while Original-Annotator does the opposite, scoring the disfluent translations as hypotheses against the cleaned references.  
We see that scoring disfluent data against clean references has a greater impact on BLEU than the opposite: the Original-Annotator BLEU is much lower, demonstrating the  significant impact that disfluent outputs can have when scoring translations of an MT system expecting clean output. 
We later use these scores as benchmarks for different training data conditions.
For all values in Table \ref{annotator agreement bleu} variance is less than 0.25 BLEU.

\begin{table}[ht]
\centering
\begin{tabular}{l|c|c|c} \hline
\bf Comparison & \texttt{dev}  & \texttt{dev2} & \texttt{test} \\ \hline 
MTurk Inter-Annotator BLEU           & 63.04 & 64.32 & 64.00 \\
Original Inter-Annotator BLEU        & 34.81 & 35.80 & 33.85 \\ \hline
Annotator-Original BLEU & 28.45 & 28.90 & 28.31 \\
Original-Annotator BLEU & 21.00 & 21.44 & 20.82 \\ \hline
\end{tabular}
\caption{Measures of MTurk annotator agreement: Inter-Annotator BLEU between generated MTurk translations, and among the 4 original translations. For comparison, BLEU between the clean references to the original disfluent refs.}
\label{annotator agreement bleu}
\end{table}

We further look at a data sample to verify the content changed is disfluent.
We find a set of 268 unique filler words within the original translations after punctuation is removed, 119 ignoring case, in part because they were crowdsourced and different translators used slightly different schemes. 
Of the tokens deleted by Turkers, 9.5\% are filler words. 
Specifically for \texttt{dev}, we find Turkers disagree on at least one token in 56\% of utterances.
Of these, some involve context-dependent disfluencies such as backchanneling, or corrections where Turkers re-phrased with minor differences. 
In most cases, backchanneling was marked as disfluent and reduced to `None').
Some disagreements involve insertions, either of pronouns (e.g. `imagined it' $\rightarrow$ `i imagined it'), or function words to make utterances more grammatical in English where Turkers introduced different tokens.
A larger percentage of disagreements involved deletion of transitional words or phrases (sentence-initial `and') to make utterances more sentence-like. In these cases,  Turkers typically removed overlapping spans,
with disagreements based on the span of tokens removed.

\section{Experiments}
\label{sec:exp}

Initial work on the original Fisher-Spanish dataset used traditional HMM-GMM ASR systems chained together with phrase-based MT systems using lattices \cite{post2013improved,kumar2014some}.
More recently, it was demonstrated in \cite{weiss2017sequence} that end-to-end sequence-to-sequence models perform competitively on this task. 

We here focus on translation from the Spanish text transcripts as an initial exploration of the problem of translating directly from noisy speech to clean references without a separate disfluency removal step.
We use sequence-to-sequence models and, as a baseline, first demonstrate the efficacy of our models on the original disfluent Fisher Spanish-English task, comparing to the previously reported numbers on the MT subtask \cite{post2013improved,kumar2014some,weiss2017sequence}.
Post et al. \cite{post2013improved} and Kumar et al. \cite{kumar2014some} are both traditional systems, while Weiss et al. \cite{weiss2017sequence} is a deep LSTM-based sequence-to-sequence model.
We then compare these results with models trained using our collected clean target data.
Finally, we look at the mismatched case where we train on disfluent data and evaluate on a cleaned test set; this is a more realistic scenario, as clean training data is difficult to collect, and we cannot expect to have it for each language and use case we encounter.
We hope this will spur future work on this lower-resource task.

We compare LSTM-based models, similar to \cite{weiss2017sequence}, to Transformer \cite{vaswani2017attention} models as implemented in OpenNMT \cite{opennmt}.
Our LSTM models use a two-layer bidirectional LSTM encoder and two-layer LSTM decoder, 500-dim embeddings, and Luong attention \cite{Luong2015b}.
We follow the default OpenNMT training procedure, optimizing with SGD for 13 epochs using a batch size of 32.
Our Transformer models follow the suggested parameters from OpenNMT, with layer size 512, sinusoidal position encodings, dropout of 0.1, label smoothing set to 0.1 \cite{szegedy2016rethinking}, and optimizing with adam using the suggested learning rate scheme. 
We reduce the number of layers to four for our smaller dataset.
We batch and normalize by tokens, and compute gradients based on four batches. We experimented with four batch sizes holding other parameters constant \{548,1096,1644,2192\}, and determined 1644 is the best for this dataset; all reported numbers use this value.
All models use the same preprocessing as previous work on this dataset \cite{post2013improved,kumar2014some,weiss2017sequence}: lowercasing and removing punctuation.

\begin{table}[ht]
\centering
\setlength\tabcolsep{4pt} 
\begin{tabular}{l|cc|cc|cc} \hline
\bf             & \multicolumn{2}{c|}{\bf dev} & \multicolumn{2}{c|}{\bf dev2} & \multicolumn{2}{c}{\bf test} \\ 
\bf System      & \bf 1R & \bf 4R & \bf 1R & \bf 4R & \bf 1R & \bf 4R \\ \hline
\bf LSTM        & 35.2 & 61.9 & 36.3 & 62.8 & 33.3 & 60.4 \\
\bf Transformer & 32.1 & 57.0 & 32.7 & 58.1 & 30.6  & 55.4 \\ \hline 
\bf Post et al. \cite{post2013improved} & -- & -- & -- & -- & -- & 58.7 \\ 
\bf Kumar et al. \cite{kumar2014some} & -- & -- & -- & 65.4 & -- & 62.9 \\ 
\bf Weiss et al. \cite{weiss2017sequence} & -- & 58.7 & -- & 59.9 & -- & 57.9 \\ 
\hline
\end{tabular}
\caption{BLEU score using \textbf{original disfluent references}. Comparing average single reference score (1R) vs multi-reference score using all four references (4R).}
\label{original references bleu}
\end{table}

Table \ref{original references bleu} shows our results on the original disfluent data. 
We provide both single and multi-reference scores: Fisher has four reference translations for \texttt{dev}, \texttt{dev2}, and \texttt{test}, which boosts scores considerably as hypotheses can match in any of the references. 
We do not have four references for our clean data, so the single reference scores provide a better basis for comparison to the clean target task.
We show both of our models perform competitively, approaching or exceeding previous best results.
Further, our single reference scores approach the inner-annotator BLEU between the four human-generated references, shown in Table \ref{annotator agreement bleu}.
For \texttt{test}, our LSTM model has a BLEU of 33.3 on a single reference, as compared to 33.8 between the four human translators.
The LSTM model is consistently slightly better than the Transformer model.
We note though that the Transformer is quite sensitive; it is possible with other parameters, it would perform better.

\begin{table}[ht]
\centering
\setlength\tabcolsep{4pt} 
\begin{tabular}{l|cc|cc|cc} \hline
\bf             & \multicolumn{2}{c|}{\bf dev} & \multicolumn{2}{c|}{\bf dev2} & \multicolumn{2}{c}{\bf test} \\ 
\bf System      & \bf 1R & \bf 2R & \bf 1R & \bf 2R & \bf 1R & \bf 2R \\ \hline
\bf LSTM        & 28.18 & 34.07 & 28.87 & 35.44 & 27.96 & 33.84 \\ 
\bf Transformer & 26.20 & 32.16 & 27.27 & 33.87 & 26.31 & 31.89 \\ 
\hline
\end{tabular}
\caption{BLEU score using \textbf{new cleaned references} to train and evaluate. Comparing average single reference score (1R) vs multi-reference score using both generated references (2R).}
\label{cleaned references bleu}
\end{table}

Turning to the task of generating clean translations, we now make use of our clean target data to train. 
Table \ref{cleaned references bleu} shows our results using this new data. 
BLEU scores go down on the clean task; a main contributor is filler words, which previously may have been overgenerated and provided partial n-gram matches and have now been removed.
For example, removing only our list of 119 filler words from the original references and scoring our disfluent LSTM model's with a single reference drops the score on \texttt{test} from 33.8 to 18.40.
In this dataset, filler words are quite one-to-one and easy to generate (see Table \ref{example disfluencies}).
Our systems improve on the mismatched condition, learning not to generate some disfluencies: we see scores on average 5.5 BLEU higher than the Original-Annotator scores in Table \ref{annotator agreement bleu}, which scores the disfluent target data against our new clean references, and can be seen as a lower bound.
Further, the original Spanish-English data is mostly one-to-one and monotonic. 
With the cleaned targets, the alignment between source and target is not as clear, making the translation task harder. 
Finally, the utterances, which were quite short to begin with, are now three tokens shorter on average.
This means a single mistake has higher consequences for BLEU, which uses 4-gram precision.

\begin{figure}[ht]
  \vspace{-0.5em}
  \centering
  \includegraphics[width=0.8\linewidth]{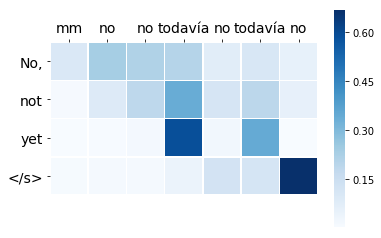}
  \vspace{-1em}
  \caption{LSTM attention: with cleaned target data, learns to place less weight on source disfluencies}
  \label{attnviz}
\end{figure}

Both architectures are able to learn to remove many disfluencies using the clean target data.
Figure \ref{attnviz} shows an example from the LSTM model attention where it has clearly learned to place less weight on source disfluencies, generating the fluent translation `No not yet.' 
Here the LSTM model has learned to both delete filler words (`mm') and repeated words and phrases (`no no', `no todavía').

Table \ref{lstm transformer output example} shows an example of the different translations generated by the LSTM and Transformer models.

\begin{table}[ht]
\centering
\setlength\tabcolsep{3pt} 
\begin{tabular}{ll}
\bf SRC  & también tengo um eh estoy tomando una clase ... \\
\bf REF  & i also have um eh im taking a marketing class ... \\
\bf CLEAN & im taking a marketing class ... \\ \hline
\bf LSTM & im taking a class of marketing  \\
\bf Transformer & i also have a class of marketing classes
\end{tabular}
\caption{Example outputs training with clean target data}
\label{lstm transformer output example}
\end{table}

While we make use of cleaned target references here, these references are expensive and time-consuming to create, even making use of crowdsourcing. We cannot expect to have this resource to train on for every language and use case, though we may have smaller datasets available for dev and test. 
Simulating this more likely scenario, Table \ref{no training data bleu} shows our results training on the original parallel disfluent data, and evaluating on cleaned dev and test data. 
As expected, we do similarly to the Annotator-Original scores in Table \ref{annotator agreement bleu}.
Using clean data, the entropy after each word is much lower, leading to lower perplexities and clearer, more concise translations. 
With the disfluent data, the model is less able to transition between meaningful tokens: filler words and clause restarts are able to appear in many places, causing the model to stutter.
It not only generates filler words and repetitions, but loses general coherence, ex) \textit{`I would tell you, I mean, it's more, it's easier, no, I mean'.}
In general, disfluent models overgenerate, producing utterances $1.25\times$ longer than our clean models. 

\begin{table}[ht]
\centering
\setlength\tabcolsep{4pt} 
\begin{tabular}{l|cc|cc|cc} \hline
\bf             & \multicolumn{2}{c|}{\bf dev} & \multicolumn{2}{c|}{\bf dev2} & \multicolumn{2}{c}{\bf test} \\ 
\bf System      & \bf 1R & \bf 2R & \bf 1R & \bf 2R & \bf 1R & \bf 2R \\ \hline
\bf LSTM        & 20.88 & 26.11 & 22.03 & 27.58 & 20.68 & 26.01 \\
\bf Transformer & 19.50 & 24.35 & 21.52 & 26.48 & 20.52 & 25.72 \\ 
\hline
\end{tabular}
\caption{\textbf{No cleaned training data condition}: BLEU score training on disfluent target data and evaluating on cleaned references. Comparing average single reference score (1R) vs multi-reference score using both generated references (2R).}
\label{no training data bleu}
\end{table}

Though the Transformer here is consistently slightly worse than the LSTM models, we speculate it could more easily be extended to lower-training data settings.
Self-attention within the Transformer models' decoder allow the decoder to attend to all previous decoder timestamps \cite{vaswani2017attention}.
We hypothesize that this mechanism could help the decoder better learn to generate clean fluent text from clean training data, particularly when disfluencies depend on the generated context (corrections, etc).
By attending to previous decoder states, the model may learn not to generate repeated words, for example. 
Comparing the LSTM and Transformer models trained with the clean target data, we see this borne out: the Transformer model has two-thirds fewer generated repetitions on \texttt{test}. 
However, both architectures learn to remove repeated words quite well: the original \texttt{test} references contain 657 repeated tokens, while and the LSTM model generates only 67, and the Transformer 44. 
We will investigate this claim in future work by pre-training this model on more fluent parallel data, such as TED, to see if pre-training the decoder self-attention enables us to do this task without requiring as much cleaned training data.
We hope that this data and our initial results encourage further work on this task, and additionally provide a benchmark to aim for using less clean target data to train, the more common condition in conversational speech translation.

\section{Conclusion}
Machine translation systems for speech suffer due to conversational speech phenomena, particularly the presence of disfluencies.
Removing disfluencies improves performance of downstream translation, as it causes data to better match typically clean training text.
However, previous work on removing disfluencies for speech translation have done so as a separate step in between speech recognition and machine translation, which is not possible using end-to-end systems. 
We release cleaned target data for a parallel speech and text corpus, enabling further work in this area. 
We compare disfluency handling among two architectures, and present a baseline to implicitly remove disfluencies within the context of end-to-end translation models.

\bibliography{bibliography}
\bibliographystyle{IEEEbib.bst}
\end{document}